\documentclass{article}
\usepackage{spconf,amsmath,graphicx}
\usepackage{amsfonts,amssymb} 
\usepackage{multirow}
\usepackage{mathrsfs}

\title{Look, Investigate, and Classify: A Deep Hybrid Attention Method for Breast Cancer Classification}
\name{\begin{tabular}{c}Bolei Xu$^{\star}$, Jingxin Liu$^{\star}$, Xianxu Hou$^{\star}$, Bozhi Liu$^{\star}$, Jon Garibaldi$^{\dagger}$, Ian O. Ellis $^{\dagger}$, \\Andy Green $^{\dagger}$, Linlin Shen$^{\star}$, Guoping Qiu$^{\star \dagger}$\end{tabular}}

\address{$^{\star}$ Shenzhen University, Shenzhen, China \\
     $^{\dagger}$ University of Nottingham, Nottingham, United Kingdom}
\begin{document}
%\ninept
%
\maketitle
\begin{abstract}
One issue with computer based histopathology image analysis is that the size of the raw image is usually very large. Taking the raw image as input to the deep learning model would be computationally expensive while resizing the raw image to low resolution would incur information loss. In this paper, we present a novel deep hybrid attention approach to breast cancer classification. It first adaptively selects a sequence of coarse regions from the raw image by a hard visual attention algorithm, and then for each such region it is able to investigate the abnormal parts based on a soft-attention mechanism. A recurrent network is then built to make decisions to classify the image region and also to predict the location of the image region to be investigated at the next time step. As the region selection process is non-differentiable, we optimize the whole network through a reinforcement approach to learn an optimal policy to classify the regions. Based on this novel Look, Investigate and Classify approach, we only need to process a fraction of the pixels in the raw image resulting in significant saving in computational resources without sacrificing performances. Our approach is evaluated on a public breast cancer histopathology database, where it demonstrates superior performance to the state-of-the-art deep learning approaches, achieving around 96\% classification accuracy while only 15\% of raw pixels are used.
\end{abstract}
\section{Introduction}
\label{sec:intro}
Breast Cancer is a major concern among women for its higher mortality when comparing with other cancer death \cite{american2008cancer}. Thus, early detection and accurate assessment are necessary to increase survival rates. In the process of clinical breast examination, it is usually fatigue and time-consuming to obtain diagnostic report by pathologist. Thus, there is large demand to develop computer-aided diagnosis (CADx) to relieve workload from pathologists.

In recent years, deep learning approaches are widely applied to the histopathology image analysis for its significant performance on various medical imaging tasks. However, one issue with deep learning approaches is that the size of raw image is large. By directly inputting raw images to the deep neural network, it would be computational expensive and requires days to train on GPUs. Some previous approaches address this problem by either resizing raw images to low resolution \cite{spanhol2017deep,han2017breast,motlagh2018breast} or randomly cropping patches \cite{rakhlin2018deep} from raw images. However, both approaches would lead to information loss and the detailed features of abnormality part could be missing, which might cause the misdiagnosed result. Another approach is to use sliding-window to crop image patches. However, there would be a large number of patches that are not related to the lesion part, since in some cases the abnormality part is usually in small portion. 

One property of human visual system is that it does not have to process the whole image at once. In clinical diagnose, pathologist would first selectively pay attention to the abnormality region, and then investigate the region for details. In this paper, we formulate the problem as a Partially Observed Markov Decision Process \cite{mnih2014recurrent}, and we propose a novel deep hybrid attention model to mimic human perception system. We build a recurrent model that is able to select image patches that are highly related to abnormality part from raw image at each time step, which so-called the ``hard-attention". Instead of directly working on the raw image, we could thus learn image features from the cropped patch. We further investigate the cropped patch through a ``soft-attention" mechanism that is to highlight pixels most related to the lesion part for classification. It should be noticed that our approach does not directly access to the raw image, and thus the computation amount of our approach is \textit{independent of the raw image size}. The patch selection process is non-differentiable, we regard the problem as a control problem, and thus could optimize the network through a reinforcement learning approach.

\begin{figure*}[htbp]
	\centering
	{\includegraphics[width=6in]{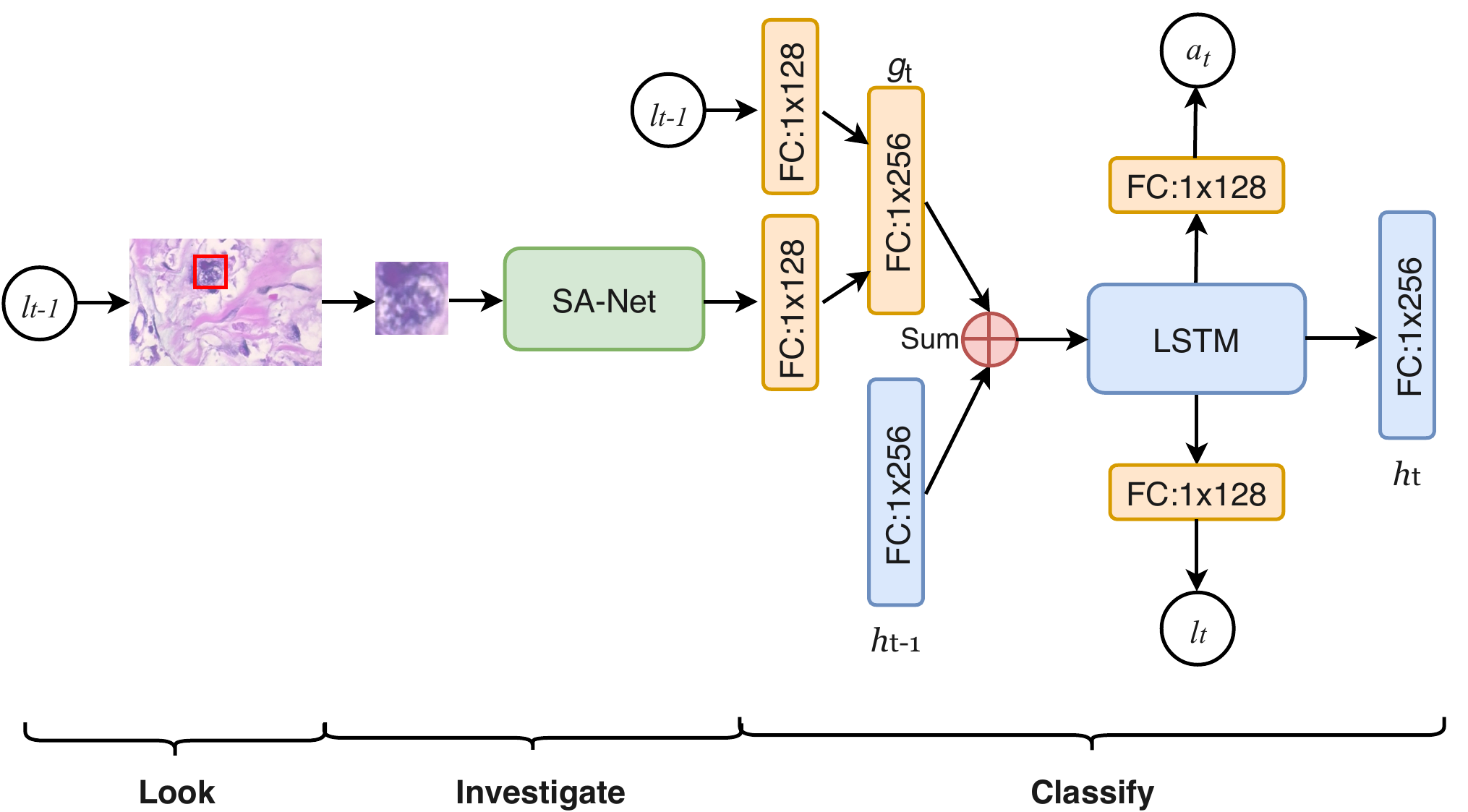}}
	\caption{The overall framework of our deep hybrid attention network. "FC" denotes fully-connected layer with ReLu activation. In each time step, the network has three stage to classify image. In the ``Look" stage, a patch is cropped by hard-attention. Then in the ``Investigate" stage, the abnormal features of image patch are extracted by the SA-Net as shown in Figure. \ref{sanet}. Finally, in the ``Classify" stage, a LSTM is employed to process the image features and also to classify image and to predict region for the next time step. For each raw image. the network crops five patches for classification.}
	\label{framework}
\end{figure*} 
The contribution of this paper could be summarized in three-fold: (1) A novel framework is introduced to the classification of breast cancer histopathology image based on the hybrid attention mechanism. (2) The proposed approach can automatically select useful region from raw image, which is able to prevent information loss and also to save computational cost. (3) Our approach demonstrates superior performance to previous state-of-the-art methods on a public dataset.

\section{Methodology}

\subsection{Network Architecture}

We formulate the histopathology image classification problem as a Partially Observable Markov Decision Process (POMDP), which means at each time step, the network does not have full access to the image and it has to make decisions based on the current observed region. It takes three stages including ``Look", ``Investigate" and ``Classify" stages as shown in Figure. \ref{framework}.

\textbf{Look Stage:} %The deep hybrid attention network is consisting of both hard and soft attention mechanisms, and also an encoder to extract \textit{glimpse} features. 
At each time step $t$, a \textit{hard-attention sensor} receives a partial image patch $x_{t}$ based on the location information $l_{t-1}$, which has smaller image size than the raw image $\text{x}$. It is a coarse region that might be related to abnormality part.

\begin{figure*}[htbp]
 	\centering
 	{\includegraphics[width=5.5in]{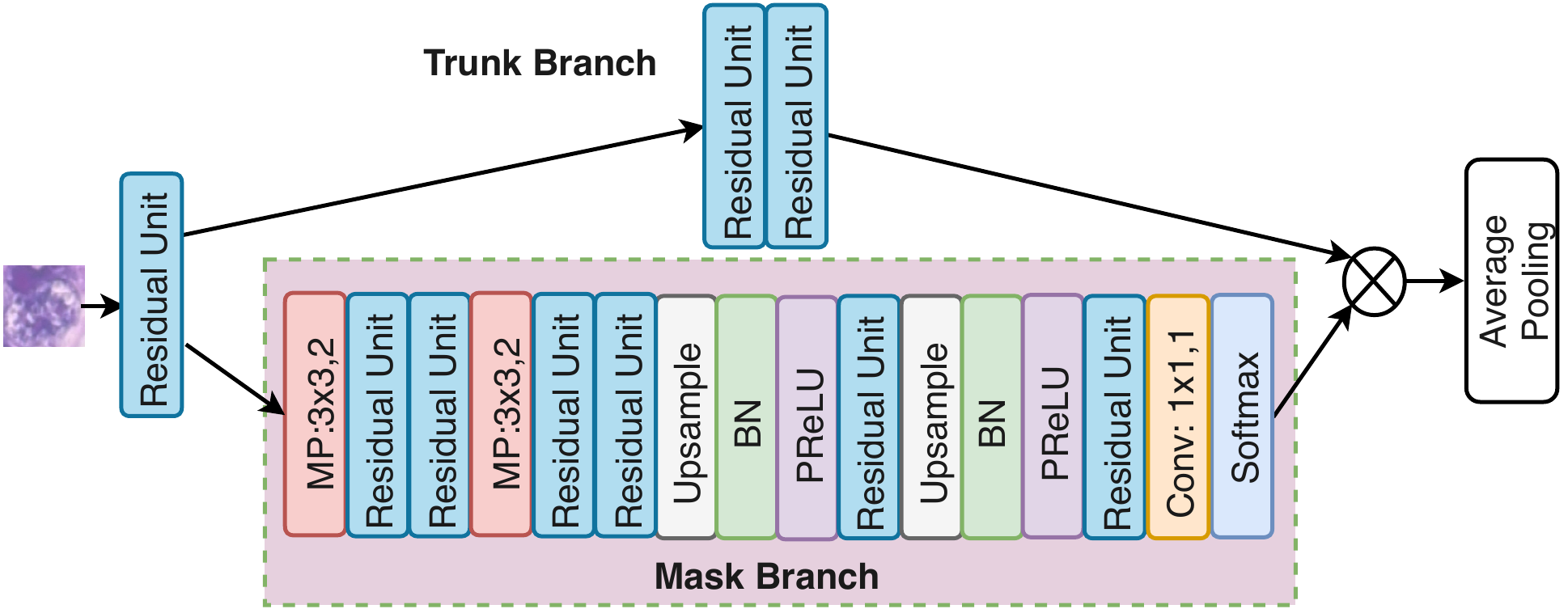}}
 	\caption{The structure of SA-Net. Here Conv($1\times 1,1$) denotes a convolutional layer with kernel size of $1$ and stride of $1$. We use 64 convolutional filters for the last Conv layers. 'BN' denotes batch normalization. MP($3\times 3,2$) means max-pooling size is set to 3 and stride is 2. 'PReLU' refers to the activation function PReLU is applied. 'Upsample' denotes upsampling by bilinear interpolation. The sturcture of residual unit is shown in Figure.\ref{res}}
 	\label{sanet}
 \end{figure*}

\textbf{Investigate Stage:} The \textit{soft-attention} mechanism $f_{s}(x_{t};\theta_{f})$ that is parameterized by $\theta_{f}$ encodes the observed image region $x_{t}$ to a soft-attention map where the valuable information is highlighted. It is achieved by a soft-attention network (SA-Net) as shown in Figure.\ref{sanet}. In the SA-Net, it contains a mask branch and a trunk branch. The soft mask branch aims to learn a mask $M(x_{t})$ in range of $[0,1]$ by a symmetrical top-down architecture and a sigmoid layer to normalize the output. The trunk branch outputs the feature map $T(x_{t})$ and the final attention map is computed by:
\begin{equation}
	\mathcal{A}(x_{t})=(1+M(x_{t}))*T(x_{t}),
\end{equation}
and the soft-attention features $f_{s}(x_{t};\theta_{f})$  are then learned by a global average pooling over the attention map $\mathcal{A}(x_{t})$. In order to fuse both learned attention features and location information, we build a fusion network
 $\mathcal{H}$ to finally produce fused feature vector $g_{t}=\mathcal{H}(f_{s}(x_{t};\theta_{f}),l_{t-1};\theta_{g})$ based on a fully-connected layer with ReLu activation.
 \begin{figure}[htbp]
 	\centering
 	{\includegraphics[width=0.15\textwidth]{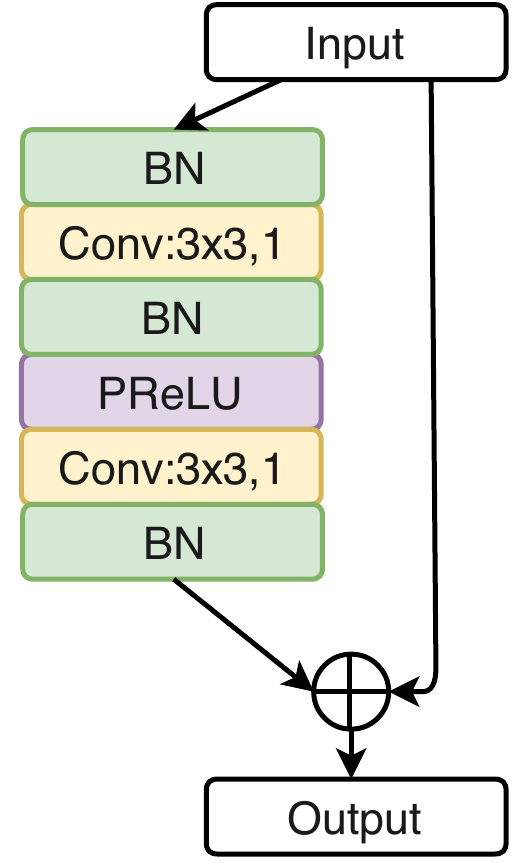}}
 	\caption{The structure of residual unit in SA-Net. We use 64 convolutional filters in each Conv layer.}
 	\label{res}
 \end{figure}

\textbf{Classify Stage:} We further use a LSTM to process the learned fused feature $g_{t}$. The advantage of LSTM is that it is able to summarize the past information, and to learn an optimal classification policy $\pi((l_{t},a_{t})|s_{1:t};\theta)$, where $a_{t}$ is decision to classify image at time step t and $s_{1:t}$ represents the past history $s_{1:t}=x_{1},l_{1},a_{1},...,x_{t-1},l_{t-1},a_{t-1},x_{t}$. The internal state is formed and updated by the hidden unit $h_{t}$ in LSTM \cite{hochreiter1997long}: $h_{t}=f_{h}(h_{t-1},g_{t};\theta_{h})$.
The recurrent LSTM network then has to choose actions including how to classify image and where to look at in the next time step based on the internal state. In this work, both actions are drawn stochastically from two distributions. The classification action $a_{t}$ is drawn from classification network by softmax output at step $t$: $a_{t}\sim(\cdot|f_{a}(h_{t};\theta_{a}))$. Similarly, the location $l_{t}$ is also drawn from a location network by $l_{t}\sim(\cdot|f_{l}(h_{t};\theta_{l}))$.

When executing the chosen actions, we could receive a image patch $x_{t+1}$ and also a \textit{reward} $r_{t+1}$ referring to whether we have correctly classified image. The total reward could be written as: $R=\sum_{t=1}^{T}r_{t}$. In this paper, we set reward to 0 for all other time steps except the last time step. In the last time step, the reward is set to 1 if the image is classified correctly and 0 if not. 

\subsection{Network Optimization}

As the hard-attention mechanism is non-differentiable, we optimize the whole network through policy gradient approach. In this paper, we aim to maximize the reward as:
\begin{equation}
	J(\theta)=\mathbb{E}_{p(s_{1:T};\theta)}[\sum_{t=1}^{T}r_{t}]=\mathbb{E}_{p(s_{1:T};\theta)}[R].
\end{equation}

In order to maximize $J$, the gradient of $J$ could be approximate by:
\begin{equation} \label{pg}
\begin{split}
	\nabla_{\theta}J&=\sum_{t=1}^{T}[\nabla_{\theta}\log\pi_{\theta}(a_{t},l_{t}|s_{1:t})R]\\
	&\approx\frac{1}{M}\sum_{i=1}^{M}\sum_{t=1}^{T}	\nabla_{\theta}\log\pi_{\theta}(a_{t}^{i},l_{t}^{i}|s_{1:t}^{i})R^{i}
\end{split}	
\end{equation}
where $i=1...M$ is the running epochs \cite{williams1992simple}. Equation. \ref{pg} encourages network to adjust parameters for the chosen probability of actions that would lead to high cumulative reward and to decrease probability of actions that would decrease reward. To achieve this, we could update the network by:
\begin{equation}\label{update}
	\theta\leftarrow\theta+\alpha\nabla_{\theta}J(\theta).
\end{equation}
 
At the meanwhile, we could also combine Equation. \ref{update} with the supervised classification training approach, i.e. to also train the network by the cross-entropy loss with ground-truth label. Thus, the network could be learned by minimizing the total loss:
\begin{equation}\label{update2}
\mathcal{L}_{total}=-J(\theta)+\mathcal{L}_{c}(y,\hat{y}),
\end{equation}
where $y$ is the ground-truth classification label, $\hat{y}$ is predicted label from network, and $\mathcal{L}_{c}$ is the cross-entropy classification loss.

\section{Experiment}
\subsection{Datasets and Parameters Setting}
We evaluated our approach on a public dataset BreakHis \cite{spanhol2016dataset}. The dataset contains 7,909 images collected from 82 patients including 58 for malignant and 24 for benign. These tumor tissue images are captured at four kinds of optical magnifications of $40\times$,  $100\times $, $200 \times$ , and $400\times$.

In the experiment, we randomly select 58 patients (70\%) for training and 24 patients (30\%) for testing. Before training, we augmented raw image by applying rotation, horizontal and vertical flips, which results in 3 times the original training data. The raw image size in the dataset is $740\times 460$. The size of five cropped images in our network is set to $112\times112$, which means we only have to process around 15\% pixels of raw image. We choose Adam optimizer with a learning rate of $0.01$ that exponentially decay over epochs. In the training stage, it usually takes around 200 epochs to convergence. The experiment is conducted on a workstation with four Nvidia 1080 Ti GPUs.

The performance of our approach is evaluated by the Patient recognition rate (PRR), in order to be comparable with previous work. PRR aims to calculate a ratio of correctly classified tissues to all the number of tissues. It could be formulated as:
\begin{equation}
	PRR = \frac{\sum_{i=1}^{N}ACC_{i}}{N},ACC=\frac{N_{rec}}{N_{p}}
\end{equation}
where $N$ is the total number of patients in the testing data. $N_{rec}$ is the correctly classified tissues of patient $p$ and $N_{p}$ is total tissue number from patient  $p$.

\subsection{Comparison with other approaches}
\begin{table}[]
	\centering
	\scriptsize

	\caption{Performance comparison of magnification specific system (in \%).``Ours w/o SA" denotes the SA-Net is removed. n/a denotes the authors did not report the corresponding data.}
	\begin{tabular}{|l|c|c|c|c|}
		\hline
		\multicolumn{1}{|c|}{\multirow{2}{*}{Methods}} & \multicolumn{4}{c|}{Magnification}                            \\ \cline{2-5} 
		\multicolumn{1}{|c|}{}                         & $40\times$            & $100\times $         & $200 \times$          & $400\times $          \\ \hline
		Spanhol \cite{spanhol2016dataset}                                       & $83.8\pm4.1$          & $82.1\pm4.9$          & $85.1\pm3.1$          & $82.3\pm3.8$          \\ \hline
		Spanhol \cite{spanhol2016breast}                                       & $90.0\pm6.7$          & $88.4\pm4.8$          & $84.6\pm4.2$          & $86.1\pm6.2 $         \\ \hline
		Gupta   \cite{gupta2017breast}                                       & $86.7\pm2.3$          & $88.6\pm2.7$          & $90.3\pm3.7$          & $88.3\pm3.0$          \\ \hline
		Sequential \cite{gupta2018sequential}                                    & $94.7\pm0.8$          & $95.9\pm4.2$          & $96.7\pm1.1$          & $89.1\pm0.1$          \\ \hline
		FV+CNN  \cite{song2017adapting}                                       & $90.0\pm3.2$          & $88.9\pm5.0$          & $86.9\pm5.2$          & $86.3\pm7.0$          \\ \hline
		MIL+CNN \cite{wu2015deep}                                       & $81.3\pm$ n/a         & $80.4\pm$  n/a          & $77.6\pm$  n/a         & $79.1\pm$  n/a         \\ \hline
		MIL    \cite{das2018multiple}                                        & $89.5\pm$ n/a          & $89.0\pm$  n/a         & $88.8\pm$  n/a        & $87.7\pm$  n/a         \\ \hline
		S-CNN  \cite{han2017breast}                               & $94.1\pm 2.1$          & $93.2\pm 1.4$         & $94.7\pm 3.6$          & $93.5\pm 2.7$         \\ \hline
		Ours w/o SA                              & $88.6\pm 1.9$          & $87.0\pm 1.8$          & $86.6\pm2.8$          & $85.2\pm1.9$          \\ \hline
		Ours                                           & $97.5\pm 1.6$ & $96.2\pm 1.3$ & $97.4\pm 2.5$ & $95.4\pm 1.5$ \\ \hline
	\end{tabular}
		\label{exp:result}
\end{table}

 \begin{figure}[htbp]
	\centering
	{\includegraphics[width=0.48\textwidth]{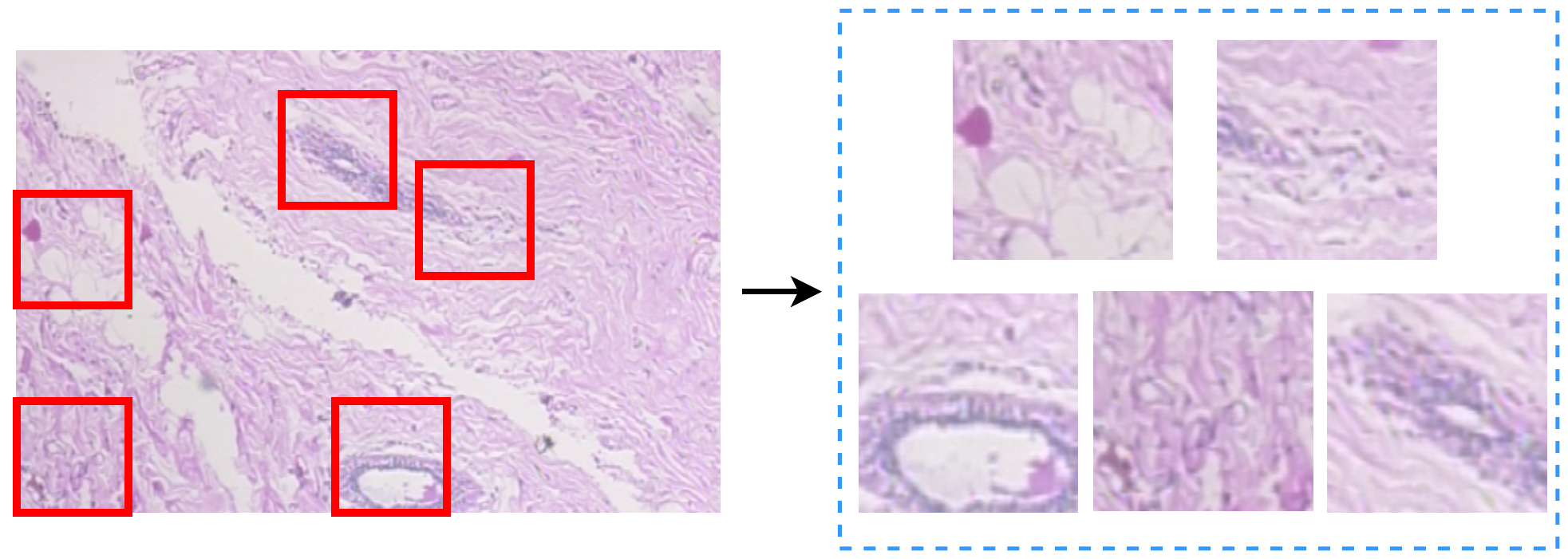}}
	\caption{An example of how hard-attention mechanism selects image patches.}
	\label{patch}
\end{figure}

To evaluate the performance of our approach to histopathology image classification, we compare our proposed deep learning framework with the state-of-the-art approaches. The results is shown in Table.\ref{exp:result} which demonstrates our approach outperforms all previous approaches. It should be noticed that our approach achieves much higher accuracy rate than most CNN approaches \cite{song2017adapting,wu2015deep,das2018multiple}. It is achieved by the well-designed attention mechanisms to select useful regions for the decision network (Figure.\ref{patch}). The hard-attention mechanism finds out the regions most related to abnormality part and the soft-attention mechanism highlight those abnormal features. 
Apart from the superior performance to the previous approaches, our approaches prevents to resize raw image which might leads to information loss, and also enables network to process image in the small size image patch in order to save computational cost.

We also conducted an ablation study to evaluate the effectiveness of the soft-attention. We remove SA-Net to test the performance of rest network. It could be seen that classification accuracy dropped down by around 10\%. The decreasing of performance is due to some redundant features are also processed by the network, which might contains noise features that leading to misclassification. Thus, it is essential to apply soft-attention mechanism to highlight useful features and also encourage network to neglect those unnecessary image features.
\section{Conclusion}
In this paper, we introduce a novel deep hybrid attention network to the breast cancer histopathology image classification. The hard-attention mechanism in the network could automatically find the useful region from raw image, and thus does not have to resize raw image for the network to prevent information loss. The built-in recurrent network can make decisions to classify image and also to predict region for next time step. We evaluate our approach on a public dataset, and it achieves around 96\% accuracy on four different magnifications while only 15\% of raw image pixels are used to make decisions to classify input image.

\bibliographystyle{IEEEbib}
\bibliography{refs}

\end{document}